\documentclass{article}

\PassOptionsToPackage{numbers, compress}{natbib}

 \usepackage[preprint]{nips_2018}

\usepackage[utf8]{inputenc} 
\usepackage[T1]{fontenc}    
\usepackage{hyperref}       
\usepackage{url}            
\usepackage{booktabs}       
\usepackage{amsfonts}       
\usepackage{nicefrac}       
\usepackage{microtype}      
\usepackage[cmex10]{amsmath}
\usepackage{amssymb}
\usepackage{graphicx}
\usepackage{subfig}

\title{Insights into Fairness through Trust: Multi-scale Trust Quantification for Financial Deep Learning}

\author{
  Alexander Wong$^{1,2,3,*}$, Andrew Hryniowski$^{1,2,3}$, and Xiao Yu Wang$^{1,3}$\\
  $^{1}$ Vision and Image Processing Research Group, University of Waterloo, Waterloo, ON, Canada\\
  $^{2}$ Waterloo Artificial Intelligence Institute, University of Waterloo, Waterloo, ON, Canada\\
  $^{3}$ DarwinAI Corp., Waterloo, ON, Canada \\
  \texttt{$^{*}$ a28wong@uwaterloo.ca} \\
}

\begin{document}

\maketitle

\begin{abstract}
The success of deep learning in recent years have led to a significant increase in interest and prevalence for its adoption to tackle financial services tasks.  One particular question that often arises as a barrier to adopting deep learning for financial services is whether the developed financial deep learning models are fair in their predictions, particularly in light of strong governance and regulatory compliance requirements in the financial services industry.  A fundamental aspect of fairness that has not been explored in financial deep learning is the concept of trust, whose variations may point to an egocentric view of fairness and thus provide insights into the fairness of models.  In this study we explore the feasibility and utility of a multi-scale trust quantification strategy to gain insights into the fairness of a financial deep learning model, particularly under different scenarios at different scales.  More specifically, we conduct multi-scale trust quantification on a deep neural network for the purpose of credit card default prediction to study: 1) the overall trustworthiness of the model 2) the trust level under all possible prediction-truth relationships, 3) the trust level across the spectrum of possible predictions, 4) the trust level across different demographic groups (e.g., age, gender, and education), and 5) distribution of overall trust for an individual prediction scenario.  The insights for this proof-of-concept study demonstrate that such a multi-scale trust quantification strategy may be helpful for data scientists and regulators in financial services as part of the verification and certification of financial deep learning solutions to gain insights into fairness and trust of these solutions.
\end{abstract}

\section{Introduction}
\vspace{-0.15in}
\label{Introduction}

The success of deep learning~\cite{lecun2015deep} in recent years have led to a significant increase in interest by academic and financial communities toward the adoption of deep learning solutions for tackling financial services applications.  Given the powerful predictive modeling capabilities facilitated by deep neural network architectures, financial deep learning has been explored for a wide range of different tasks such as stock market prediction~\cite{Hsu,kumar2017opening,Singh,Chen,hu2017listening}, financial trading~\cite{cohen2019trading}, credit card fraud detection~\cite{Roy,Jurgovsky}, credit card default prediction~\cite{YEH20092473}, exchange rate prediction~\cite{Shen,Dautel}, and anti-money laundering~\cite{weber2019antimoney,4635778}.  These encouraging studies related to financial deep learning have led to financial institutions around the world to explore the deployment of such solutions in real-world financial practice.

One particular question that often arises as a barrier to adopting deep learning for financial services in real-world practice is whether the developed financial deep learning models are fair in their predictions.  Fairness in financial models is particularly crucial in light of strong governance and regulatory compliance requirements in the financial services industry, as well as the socioeconomic implications of leveraging automated decision-making tools within the global financial ecosystem.  While explorations into fairness for financial deep learning have been conducted through aspects such as explainability~\cite{kumar2017opening,modarres2018explainable}, a fundamental aspect of fairness that has not been explored in financial deep learning is the concept of trust, particularly since variations in trust under different scenarios and circumstances can indicate an egocentric view of fairness by a model, which leads to inherent biases in predictions made.  It is through this lens of trust that we try to gain insights into the fairness of financial deep learning models.

Motivated by the importance of trust in deep learning in general, there has been a new focus in recent years on trust quantification, where the goal is to quantify the trustworthiness of deep neural networks and the predictions and decisions they make.  Methods range from uncertainty estimation~\cite{titensky2018uncertainty,geifman2018biasreduced,NIPS2017_7141,gal2015dropout} to agreement assessment~\cite{jiang2018trust} to subjective logic~\cite{deeptrust}.  More recently, a number of studies have focused on the design of interpretable metrics for the purpose of trust quantification of deep neural networks at different scales~\cite{alex2020really,hryniowski2020does}.  More specifically, the trust quantification metrics introduced in these studies include (from finest to coarsest scale):

\begin{enumerate}
    \item \textbf{Question-Answer Trust}: a scalar measure of a model's trustworthiness for a single question-answer pair.\vspace{-0.05in}
    \item \textbf{Trust Density}: distribution of question-answer trust for a given prediction scenario.  An extension of trust density is the notion of conditional trust densities, which further decompose a trust density to gain insight into trust behaviour for a given prediction scenario both correct prediction and incorrect prediction scenarios.
    \item \textbf{Trust Matrix}: a matrix representation of the trust level under all possible prediction-truth relationships.
    \item \textbf{Trust Spectrum}: a measure of the overall trust in a model across spectrum of possible prediction scenarios.
    \item \textbf{NetTrustScore}: a scalar measure of overall trustworthy of a model.  An extension of NetTrustScore is the notion of conditional NetTrustScores, which further decompose NetTrustScore to gain insight into overall trust behaviour for a given model under correct prediction and incorrect prediction scenarios.

\end{enumerate}

By studying trustworthiness of a deep neural network across multiple scales, one can gain deeper insights into not just how trustworthy a deep neural network is, but also where trust breaks down.  These insights into trustworthiness can unveil important insights into the fairness of a deep neural network, particularly by observing the variations in trust under different scenarios at different scales since it may unveil an egocentric view of fairness that can lead to biases in predictions.

Motivated by the insights that can be gained with such an approach, in this study we explore the feasibility and utility of multi-scale quantification for financial deep learning to gain insights into the fairness of a model.  More specifically, we conduct proof-of-concept multi-scale trust quantification on a deep neural network for the purpose of credit card default prediction to study:
\begin{itemize}
    \item the overall trustworthiness of the model,
    \item the trust level under all possible prediction-truth relationships,
    \item the trust level across the spectrum of possible predictions,
    \item the trust level across different demographic groups (In this particular study, we study trust levels across age, gender, and education), and
    \item the distribution of overall trust for an individual prediction scenario.
\end{itemize}
We then study the trust quantification results produced by this suite of quantitative metrics and the variations observed within these metrics in great detail to gain deeper insights into the fairness of the model.  To facilitate this multi-scale trust quantification, we modify trust quantification metrics that were introduced in past studies~\cite{alex2020really,hryniowski2020does} to account for demographics in their mathematical formulation, as well as introduce new trust quantification metrics in the form of demographic trust spectra.

The paper is organized as follows.  In Section~\ref{Methods}, we discuss the multi-scale trust quantification strategy being leveraged for financial deep learning, particularly discussing the concepts of question-answer trust, trust density, trust spectrum, and NetTrustScore. We continue by modifying their mathematical formulation to account for demographics, and introduce a novel concept of demographic trust spectrum and its associated mathematical formulation.  In Section~\ref{modeldata}, we will also describe the deep neural network for credit card default prediction used to as proof-of-concept to illustrate the feasibility and utility of the multi-scale trust quantification approach, as well as the dataset used.  In Section~\ref{results}, we will discuss the trust quantification results produced by the multi-scale trust quantification process to investigate where trust breaks down for this model at different scales of granularity, and gain some valuable insights on trust and fairness for financial deep learning.

\section{Methods}
\vspace{-0.15in}
\label{Methods}
Let us first describe the methodology leveraged in this study to conduct proof-of-concept multi-scale trust quantification for financial deep learning to gain insights into the fairness of models.  More specifically, the proposed multi-scale trust quantification strategy is conducted by modifying a number of trust quantification metrics that were introduced in several studies~\cite{alex2020really,hryniowski2020does} to account for demographics in their mathematical formulation, as well as introducing new trust quantification metrics in the form of demographic trust spectra.  The motivation behind this approach is to quantify and study the trust behaviour of deep neural networks at different scales based on a set of questions to provide a more complete understanding into where trust variations lie and where trust breaks down. Knowing where trust variations lie and where trust breaks down at different scales facilitates for rapid improvements in the model through the identification of gaps in trust and fairness, as well as facilitate an additional quantitative compliment to existing certification processes.  The trust quantification metrics leveraged in this multi-scale trust quantification strategy aligns with social psychology studies~\cite{Tenney,Tenney07,Tenney08} that showed more negative sentiments and distrust towards overconfidence that is not reflected by an individual's actual performance, as well as more negative sentiments towards an individual's overcautious behaviour. This combination of metrics also aligns well with the notion that overconfidence can lead to an egocentric view of fairness, which leads to inherent biases in predictions made.  The trust quantification metrics leveraged here to gain insights on the fairness of a financial deep learning model is described below.

\subsection{Question-answer Trust}
\vspace{-0.1in}
Based on the above two premises, let us first define the relationship for an demographic-question-answer tuple $(w,x,y)$ with respect to model $M$ as
\begin{equation}
y = M(x)
\end{equation}
\noindent where $x \in X$ denotes the question, $X$ denotes the space of all possible questions, $w \in W$ denotes the demographic group (e.g., male, female, secondary school educated, university educated, between age of 20-30, etc.), $W$ denotes the space of all possible demographic groups, $y \in Z$ denotes the model answer, $z \in Z$ represents the oracle answer, and $Z$ is the space of all possible answers.  Let $R_{y \neq z|M,w}$ denote as the space of all questions $x$ belonging to demographic group $w$ where the answer $y$ by model $M$ does not match the oracle answer $z$ (i.e., incorrect answers), and $R_{y = z|M,w}$ denote the space of all questions $x$ belonging to demographic group $w$ where the answer $y$ by model $M$ matches the oracle answer $z$ (i.e., correct answers).  Furthermore, let $C(y|x)$ denote the confidence of $M$ in an answer $y$ to question $x$.  The first trust quantification metric is the question-answer trust $Q_{z,w}(x,y)$ for a given demographic-question-answer tuple $(w,x,y)$ can be expressed as
\begin{equation}
    Q_{z,w}(x,y) =
    \begin{cases}
      C(y|x)^\alpha & \text{if $x \in R_{y = z|M,w}$ } \\
      (1 - C(y|x))^\beta & \text{if $x \in R_{y \neq z|M,w}$ } \\
    \end{cases}
    \label{questiontrust}
\end{equation}
\noindent where $\alpha$ and $\beta$ denote reward and penalty relaxation coefficients.  In this study, we set $\alpha=1$ and $\beta=1$ to penalize undeserved overconfidence and overcautious behaviour equally.  The dynamic range of $Q_{z,w}(x,y)$ is from 0 to 1, where 0 indicates the lowest level of trust in the answer and 1 indicates the highest level of trust.

\subsection{Trust Density}
\vspace{-0.1in}
Based on the question-answer trust metric, the second trust quantification metric is the trust density $F(Q_{z,w})$, which can be defined as the distribution of question-answer trust $Q_{z,w}(x,y)$ across all questions $x$ that are answered as $z$.   To further decompose $F(Q_{z,w})$ based on correct and incorrect prediction scenarios, we define $F(y=z)F(Q_{z,w}|y=z)$, the conditional density of question-answer trust given a correct answer and $F(y \neq z)F(Q_{z,w}|y \neq z)$, the conditional density of question-answer trust given an incorrect answer.  The sum of the conditional densities $F(y=z)F(Q_{z,w}|y=z)$ and $F(y \neq z)F(Q_{z,w}|y \neq z)$ is equal to $F(Q_{z,w})$.  Given that the number of questions used to study the trust behaviour of a model, an approximation of the trust density is computed via non-parametric density estimation with a Gaussian kernel of bandwidth $\frac{\gamma}{\sqrt{N}}$ where $\gamma_j$ is a kernel constant. In this study, $\gamma$ is set to $0.5$, with a reflection condition is used for boundary events.

\subsection{Trust Matrix}
\vspace{-0.1in}
Based on the question-answer trust metric, the third trust quantification metric is the trust density $F(Q_{z,w})$, which can be defined as a matrix of expected question-answer trusts for every possible model-oracle answer scenario $(y,z)$.  More specifically, a single element in the trust matrix (represented here as $Q_{z}(y)$) for the unique model-oracle answer tuple $(y,z)$ can be expressed as
\begin{equation}
    Q_z(y) = E[Q_{z,w}(x,y)]
\end{equation}

\subsection{Trust Spectrum}
\vspace{-0.1in}
Furthermore, based on the question-answer trust metric, the fourth trust quantification metric is the trust spectrum, a function that characterizes the level of trust for a model with respect to the spectrum of possible answer scenarios based on question-answer trust across correctly and incorrectly answered questions. More specifically, the trust spectrum $\{T_M(z)\}_{z \in Z}$ consists of trust spectrum coefficients $T_M(z)$ across all possible answer scenarios $Z$, where $T_M(z)$ is the question-answer trust integral across all demographic-question-answer tuples $(w,x,y)$,
\begin{equation}
T_M\left(z\right) = \int \int \int P(w,x,y)Q_{z,w}(x,y) dydxdw
\label{spectrum}
\end{equation}
\noindent where $P(w,x,y)$ is the probability of the occurrence of demographic-question-answer tuple $(w,x,y)$.  The trust spectrum gives detailed insights in where trust more frequently breaks down at the answer scenario level.  The dynamic range for $T_M(z)$ is from 0 to 1, where 0 indicates the lowest level of trust in the model for a given answer scenario $z$ and 1 indicates the highest level of trust in the model for a given answer scenario $z$.

\subsection{Demographic Trust Spectra}
\vspace{-0.1in}
To further study the trust and fairness of a model with regards to different demographics, the fifth trust quantification metric is the demographic trust spectra,  functions that characterize the level of trust for a model with respect to the spectrum of possible demographic groups $W$ based on question-answer trust across answered questions. More specifically, a demographic trust spectrum $\{T_M(w)\}_{w \in W}$ consists of demographic trust spectrum coefficients $T_M(w)$ across all possible demographic groups $W$, where $T_M(w)$ is the question-answer trust integral across all question-answer pairs $(x,y)$ for a given demographic group $w$,
\begin{equation}
T_M\left(w\right) = \int \int \int P(x,y,z)Q_{z,w}(x,y) dydxdz
\label{spectrum}
\end{equation}
\noindent where $P(x,y,z)$ is the probability of the occurrence of question-answer-oracle tuple $(x,y,z)$.  The demographic trust spectrum gives detailed insights in where trust more frequently breaks down at the demographic group level.  The dynamic range for $T_M(w)$ is from 0 to 1, where 0 indicates the lowest level of trust in the model for a given demographic group  $w$ and 1 indicates the highest level of trust in the model for a given demographic group $w$.

\subsection{NetTrustScore}
\vspace{-0.1in}
Finally, the sixth trust quantification metric is NetTrustScore ($T_M$), which summarizes overall trust of a model and is expressed as the expectation of $T_M(z)$ across all possible answer scenarios $z$,

\begin{equation}
T_M = \int P(z)T_M(z) dz
\label{trustscore}
\end{equation}
\noindent where $P(z)$ is the probability of occurrence of answer scenario $z$.

The dynamic range of $T_M$ is from 0 to 1, where 0 indicates the lowest level of overall trust in the model and 1 indicates the highest overall level of trust in the model.  To further decompose $T_M$ based on correct and incorrect prediction scenarios, we define $T_{M, y=z}$, the conditional NetTrustScore given the answer is correct and $T_{M, y \neq z}$, the conditional NetTrustScore given the answer is incorrect.

\section{Model and Data}
\label{modeldata}
\vspace{-0.15in}
In this study, we conduct a proof-of-concept multi-scale trust quantification using the aforementioned suite of metrics on a deep neural network model designed for credit card default prediction to gain insights into the fairness of the model.  The task of credit card default prediction was leveraged in this proof-of-concept study because such predictive analytics models can be used to determine if an individual receives credit or not, and thus fairness and trust becomes very critical under this circumstance to avoid unfair treatment.  More specifically, the deep neural network model ($M$) consists of six fully-connected layers, with the four hidden layers having 10 neurons each.  The output layer of the deep neural network model, which represents the model answer $y$, is a two neuron softmax layer, with each output neuron representing one of two possible predictions: 1) payment default, and 2) no default.  Finally, the softmax value for a given prediction is leveraged to represent the confidence $C(y|x)$ of the given prediction.  For the set of questions $x$ used in this proof-of-concept multi-scale trust quantification, we leveraged the Taiwan credit card default dataset~\cite{YEH20092473}, which comprises of 30,000 client records with 23 exploratory variables. The dataset is balanced and the model trained with Adam optimizer for 20 epochs with a starting learning rate of 1e-3 and an exponential decay of 0.96.  It is important to note that the goal here is not to achieve state-of-the-art prediction performance, but to train a model for studying the efficacy and utility of multi-scale trust quantification for unveiling where trust breaks down in a deep neural network to provide insights into potential fairness gaps and issues that need to be addressed.

\section{Results and Discussion}
\vspace{-0.15in}
\label{results}

To explore the feasibility and utility of leveraging this suite of trust quantification metrics for multi-scale quantification for financial deep learning, we will now analyze the results for the aforementioned deep neural network model for credit card default prediction.

\begin{table}[h]
	\centering
	\caption{Accuracy, NetTrustScore ($T_M$), conditional NetTrustScore for correct answers ($T_{M, y=z}$), and conditional NetTrustScore for incorrect answers ($T_{M, y \neq z}$) for the tested deep neural network for credit card default prediction.   }
	
	\begin{tabular}{cccc}
		\hline
		Accuracy & NetTrustScore ($T_M$) & $T_{M, y=z}$ & $T_{M, y \neq z}$\\
		\hline 
		0.709 &	0.618 	&	0.734 & 0.335 \\
		\hline
	\end{tabular}\\	
	\label{tab:Results}
\end{table}	

\textbf{NetTrustScore and conditional NetTrustScores.} Table~\ref{tab:Results} shows the accuracy and NetTrustScore for the evaluated model and the corresponding conditional NetTrustScores for both correct and incorrect model answer scenarios. It can be observed from the NetTrustScore that the evaluated model achieves a moderate level of overall trustworthiness, with a level of trustworthiness that is noticeably lower than its level of accuracy.  This disparity between the overall accuracy and overall trustworthiness of the model highlights the additional insights that a trust quantification strategy can unveil about a model's behaviour beyond traditional performance metrics.  Furthermore, looking at the conditional NetTrustScores, one can see that the model achieves noticeably higher overall trustworthiness for correct answer scenarios when compared to that of incorrect answer scenarios.  What this indicates is that the model is quite overconfident about the credit card default predictions it makes, even when the actual performance is actually poor in those scenarios.

\begin{figure*}[t]
\centering
	\includegraphics[width = 0.4\linewidth]{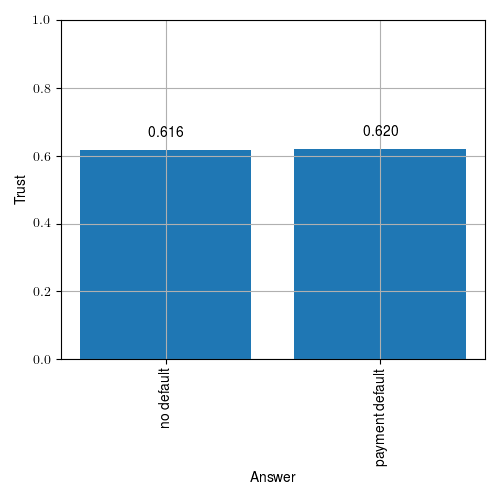}
	\caption{The trust spectrum of the deep neural network for credit card default prediction.  The model exhibits balanced, fair behaviour with respect to the types of predictions it can make given that it exhibits similar trust levels across the possible set of prediction scenarios.}
	\label{fig:trustspectrum}
\vspace{-0.2in}
\end{figure*}

\textbf{Trust Spectrum.} Next, we will study the overall level of trust for the evaluated model with respect to the spectrum of possible prediction scenarios based on its trust spectrum (see Figure~\ref{fig:trustspectrum}).  It can be observed that the expected level of trust in the evaluated model's predictions for the `no default' scenario is similar to that for the `payment default' scenario, with the trust for `no default' being slightly low than that of `payment default'.  What this indicates is that the model exhibits balanced, fair behaviour with respect to the types of predictions it can make given that it exhibits similar trust levels across the possible set of prediction scenarios.

\begin{figure}[h]
    \centering
    \begin{tabular}{ccc}
        \subfloat[Gender\label{fig:trustspectra_demographic_gender}]{\includegraphics[width=0.32\linewidth]{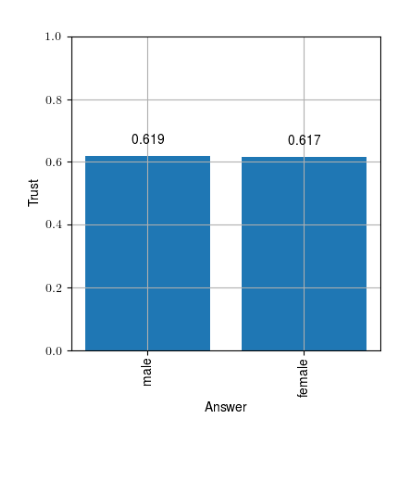}} &
        \subfloat[Education\label{fig:trustspectra_demographic_education}]{\includegraphics[width=0.32\linewidth]{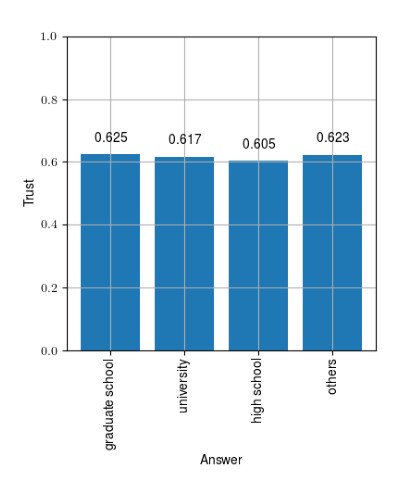}} &
        \subfloat[Age\label{fig:trustspectra_demographic_age}]{\includegraphics[width=0.32\linewidth]{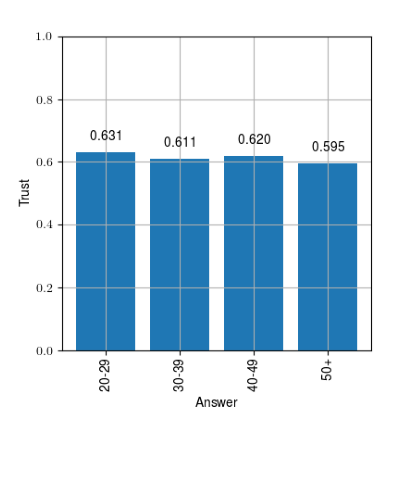}}
    \end{tabular}
    \caption{The demographic trust spectra for gender, education, and age of the deep neural network for credit card default prediction.  The model exhibits potential trust biases with respect to education and age, with warrants a deep investigation to identify mechanisms for improving fairness.}
    \label{fig:trustspectra_demographic}
\end{figure}

\textbf{Demographic Trust Spectra.} Let us now study the overall level of trust for the evaluated model with respect to the spectrum of demographic groups based on its demographic trust spectra (see Figure~\ref{fig:trustspectra_demographic}).  A number of observations can be made.  First, it can be observed from Figure~\ref{fig:trustspectra_demographic_gender} that the expected level of trust in the evaluated model's predictions are similar across both male and female demographic groups.  This indicates that there are no intrinsic biases that leads the model to be more trustworthy when making predictions for a particular gender, and thus the model is exhibiting fair behaviour in this scenario.  Second, it can be observed from Figure~\ref{fig:trustspectra_demographic_education} that the expected level of trust in the evaluated model's predictions for the population with high school education is noticeably lower when compared to the other education groups, with an expected trust gap of 2\% between those with graduate school education and those with high school education.  This observation could be indicative of unobserved characteristics associated with the high school educated demographic group that is absent from the collected data and thus worth deeper exploration to identify, as adding such characteristics could lead to a more trustworthy prediction outcomes for those with a high school education.  Third, it can be observed from Figure~\ref{fig:trustspectra_demographic_age} that the expected level of trust in the evaluated model's predictions for the age groups 30-39 and 50+ are noticeably lower when compared to the other two education groups, with the expected trustworthiness of predictions made for the age group of 50+ exhibiting a 3.6\% expected trust gap when compared to those made for the age group of 20-29.  This indicates that the model exhibits may have a potential bias against predicting trustworthy answers for those two age groups which may need to be remedied upon further investigation into the data characteristics of these demographic groups.

\begin{figure*}[t]
\centering
\vspace{-0.21 cm}
	\includegraphics[width = 0.45\linewidth]{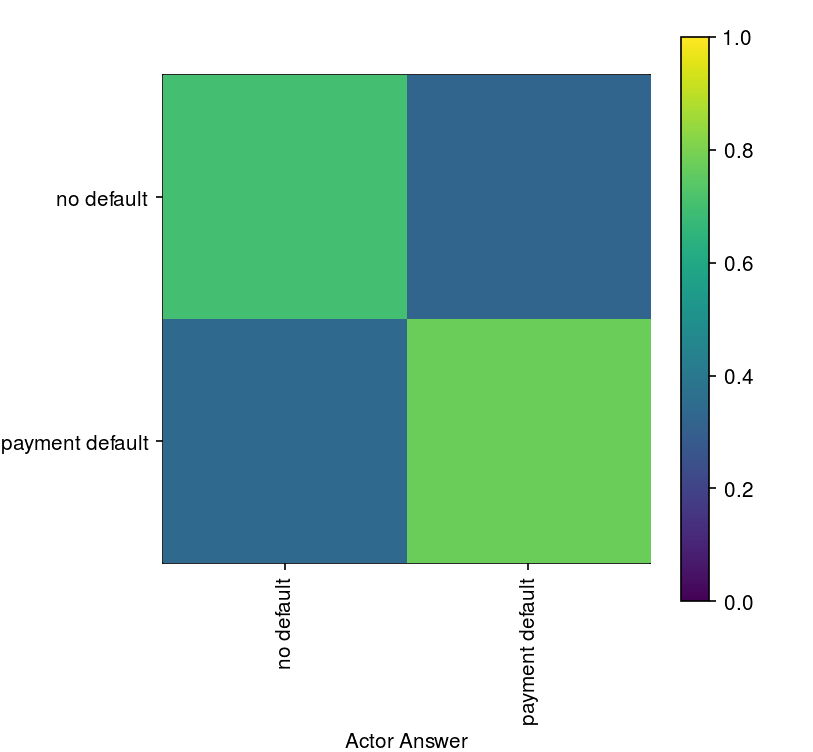}
	\caption{The trust matrix of the deep neural network for credit card default prediction.}
	\label{fig:trustmatrix}
\vspace{-0.21 cm}
\end{figure*}

\textbf{Trust Matrix.} We will now study the level of trust under all possible prediction-truth relationships based on its trust matrix (see Figure~\ref{fig:trustmatrix}).  A number of interesting observations can be seen.  First of all, it can be observed that the trust level of the model when making correct `payment default' predictions is higher than that when making correct `no default' predictions.  This indicates that the evaluated model exhibits a slight confidence bias towards `payment default'.  Second, it can be observed that the trust level of the model mistakenly predicting `payment default' when it should have been `no default' is lower than the trust level of the model mistakenly predicting `no default' when it should have been `payment default'.  This behaviour further reflects a model's confidence bias towards `payment default' predictions, where it is exhibiting more overconfidence when deciding that it is a `payment default' scenario.  These insights regarding confidence bias are useful in highlighting areas that need further improvements for greater decision fairness.

\begin{figure}[t]
    \centering
    \begin{tabular}{cc}
        \subfloat[No Default\label{fig:no_default_trustdensity_cases}]{\includegraphics[width=0.45\linewidth]{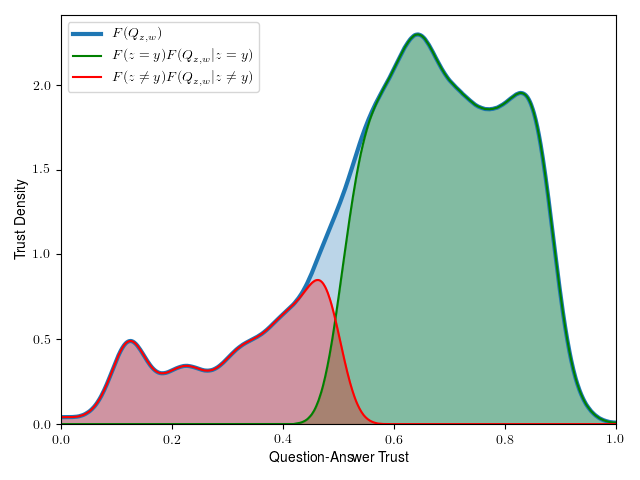}} &
        \subfloat[Payment Default\label{fig:default_trustdensity_cases}]{\includegraphics[width=0.45\linewidth]{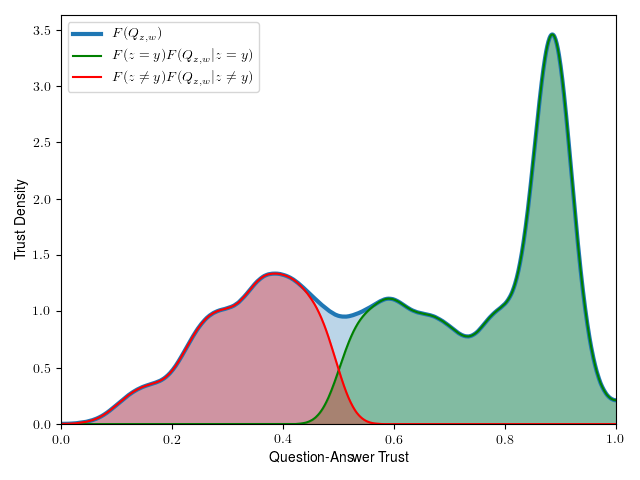}}
    \end{tabular}
    \caption{Trust densities and conditional trust densities of the deep neural network for credit card default prediction for the specific scenarios of `no default' and `payment default'.}
    \vspace{-0.15in}
    \label{fig:trustdensity_cases}
\end{figure}

\textbf{Trust Density and Conditional Trust Densities.} Finally, let us now study the distribution of trust for a given specific prediction scenario by looking at the trust density and conditional trust densities for the `no default' scenario (see Figure~\ref{fig:no_default_trustdensity_cases}) and for the `payment default' scenario (see Figure~\ref{fig:default_trustdensity_cases}).  A number of interesting observations can be seen.  First of all, it can be observed that there is a greater distribution of question-answer trust in the medium-to-high trust regions of the trust density of `no default' scenario than that of the `payment default' scenario.  On the other hand, it can be observed that there is a much higher density peak in the high trust regions of the trust density of `payment default' scenario than that of `no default' scenario.  Looking at the conditional trust densities further reveals that the high density peaks correspond to correct predictions.  These observations further reinforce the fact that when the model is very confident when right about its `payment default' predictions, and more overconfident when wrong about its `payment default' predictions than when making 'no default' predictions.  This confidence bias further illustrates room for improvement in terms of predictive fairness for the given model.

The aforementioned insights about where trust breaks down and where variations in trust exists for a given model that was unveiled in this proof-of-concept study demonstrates that a multi-scale trust quantification strategy may be helpful for data scientists and regulators in financial services as part of the verification and certification of financial deep learning solutions to gain insights into fairness and trust of these solutions.

\bibliographystyle{IEEEtran}
\bibliography{trustfinance}

\end{document}